\documentclass[letterpaper]{article} 
\usepackage{aaai2026}  
\usepackage{times}  
\usepackage{helvet}  
\usepackage{courier}  
\usepackage[hyphens]{url}  
\usepackage{graphicx} 
\usepackage{natbib}  
\usepackage{caption} 
\usepackage{algorithm}
\usepackage{algorithmic}
\usepackage{amsmath}
\usepackage{amssymb}  
\usepackage{multirow}
\usepackage{subcaption}
\usepackage{xcolor}
\usepackage{colortbl}
\usepackage{booktabs} 
\usepackage{newfloat}
\usepackage{listings}
\DeclareCaptionStyle{ruled}{labelfont=normalfont,labelsep=colon,strut=off} 
\floatstyle{ruled}
\newfloat{listing}{tb}{lst}{}
\floatname{listing}{Listing}
\title{ViSP: A PPO-Driven Framework for Sarcasm Generation with Contrastive Learning}

\title{ViSP: A PPO-Driven Framework for Sarcasm Generation with Contrastive Learning}
\author {
    Changli Wang\textsuperscript{\rm 1},
    Rui Wu \textsuperscript{\rm 1, \footnote{Corresponding author: Rui Wu.}},
    Fang Yin\textsuperscript{\rm 2}
}
\affiliations {
    \textsuperscript{\rm 1}Faculty of Computing, Harbin Institute of Technology, Harbin, China\\
    \textsuperscript{\rm 2}College of Computer Science and Technology, Harbin University of Science and Technology, Harbin, China
    simple@hit.edu.cn
}

\usepackage{bibentry}

\begin{document}

\maketitle

\begin{abstract}
Human emotions are complex, with sarcasm being a subtle and distinctive form.
Despite progress in sarcasm research, sarcasm generation remains underexplored, primarily due to the overreliance on textual modalities and the neglect of visual cues, as well as the mismatch between image content and sarcastic intent in existing datasets.
In this paper, we introduce \textbf{M2SaG}, a multimodal sarcasm generation dataset with 4,970 samples, each containing an image, a sarcastic text, and a sarcasm target.
To benchmark M2SaG, we propose ViSP, a generation framework that integrates Proximal Policy Optimization (PPO) and contrastive learning. 
PPO utilizes reward scores from DIP to steer the generation of sarcastic texts, while contrastive learning encourages the model to favor outputs with higher reward scores. 
These strategies improve overall generation quality and produce texts with more pronounced sarcastic intent.
We evaluate ViSP across five metric sets and find it surpasses all baselines, including large language models, underscoring their limitations in sarcasm generation.
Furthermore, we analyze the distributions of Sarcasm Scores and Factual Incongruity for both M2SaG and the texts generated by ViSP. 
The generated texts exhibit higher mean Sarcasm Scores (0.898 vs. 0.770) and Factual Incongruity (0.768 vs. 0.739), demonstrating that ViSP produces higher-quality sarcastic content than the original dataset.
Our dataset and code will be released at \textit{https://github.com/wclapply/ViSP}.
\end{abstract}


\section{Introduction}
Human emotions are inherently complex and multifaceted, with sarcasm representing a distinct mode of expression. 
According to Wikipedia\footnote{https://wikipedia.org/}, sarcasm is described as as a literary genre that employs rhetorical devices such as exaggeration and irony to expose contradictions or flaws, often resulting in a humorous effect.
Psychological evidence indicates that, although the use and comprehension of sarcasm require significant cognitive resources \cite{ref1}, they are positively correlated with the recipient’s ability to infer and understand the mental states of others, a capacity referred to as Theory of Mind (ToM) \cite{ref2}.
While humans effortlessly interpret multimodal cues to infer others’ mental states, AI systems face significant difficulty in this regard. 
Their lack of innate social reasoning limits their ability to grasp sarcasm, which is crucial for natural and context-sensitive human-computer interaction.

\begin{figure}[t]
\centering
	\subfloat[MuSG Dataset]{\includegraphics[width = 0.47\textwidth]{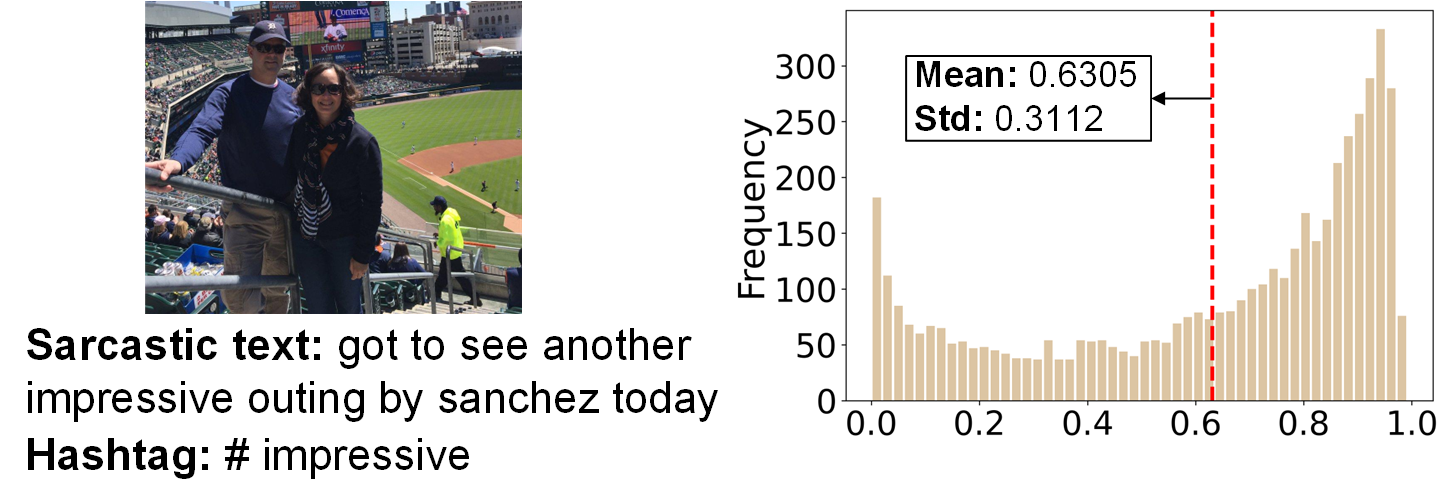}}\\
	\subfloat[M2SaG Dataset (ours)]{\includegraphics[width = 0.47\textwidth]{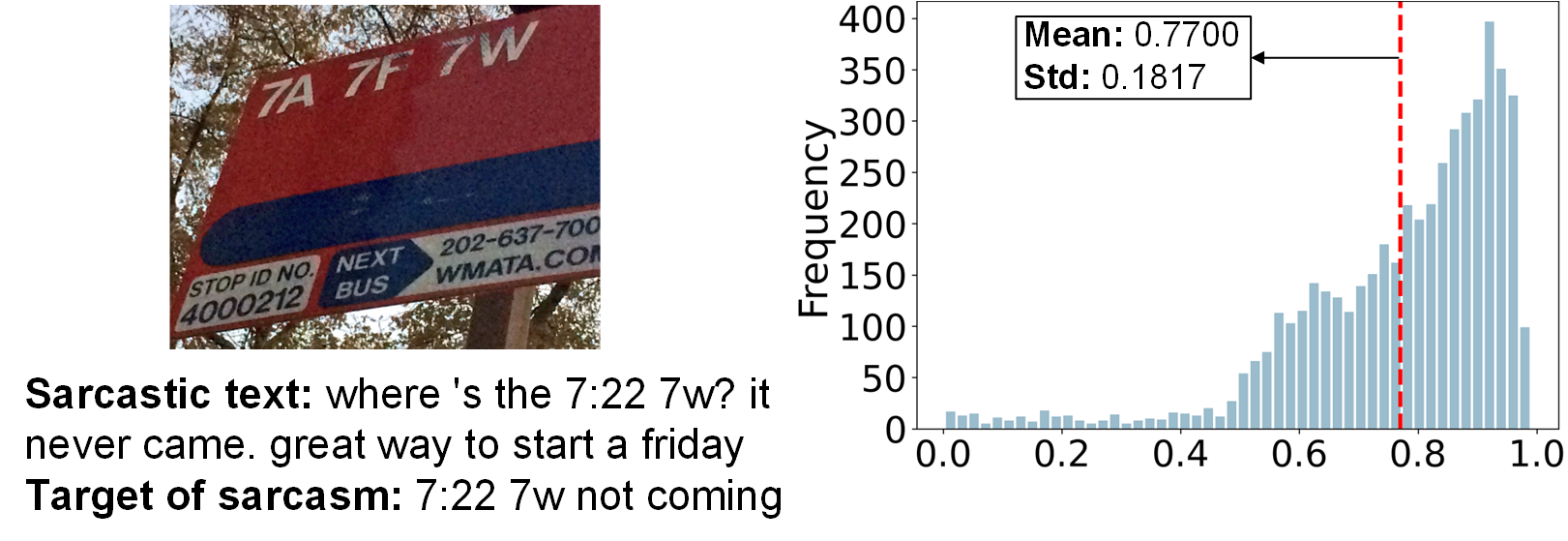}}
\caption{Comparison between MuSG and M2SaG datasets: (a) displays a MuSG sample and its sarcasm score distribution, with a mean of 0.6306 and standard deviation of 0.3112. (b) shows an M2SaG sample and its scores, achieving a higher mean of 0.7700 and lower standard deviation of 0.1817, indicating stronger and more consistent sarcasm.}
\label{fig1}
\end{figure}

\textbf{Motivation.} 
Sarcasm often manifests as a strong semantic incongruity between the image and the accompanying text, a phenomenon known as Factual Incongruity \cite{dip}.
Therefore, compared to other forms of affective text generation, multimodal sarcasm generation requires a more nuanced understanding of cross-modal semantics.
Despite notable progress in sarcasm detection and explanation, multimodal sarcasm generation remains an essential yet underexplored area within  affective computing.
We attribute the limited progress in multimodal sarcasm generation to \textbf{two principal factors.}
\textbf{First}, the majority of existing research predominantly leverages textual modalities, often neglecting the nuanced and complementary semantic cues embedded in visual content. 
This unimodal focus constrains the capacity of models to capture the full spectrum of sarcastic expression. 
\textbf{Second}, the quality of available datasets remains subpar.  
The existing sarcasm generation dataset, MuSG \cite{mtmsg}, exhibits a significant mismatch between images and sarcastic texts, as illustrated in Fig.~\ref{fig1}(a). 
To systematically evaluate sarcasm intensity, we employ DIP \cite{dip} to compute sarcasm scores for the MuSG dataset and report their mean and standard deviation (see Fig.~\ref{fig1}).
A higher score signifies a stronger sarcastic intent.
Statistically, approximately one-third of MuSG samples yield sarcasm scores below 0.5, indicating a substantial proportion of the dataset lacks explicit sarcastic expression.  
Moreover, the imprecise indication of sarcasm targets by hashtags in MuSG further impedes progress in sarcasm generation research.
For example, in the left of Fig.~\ref{fig1}(a), the hashtag \textit{\#immpressive} offers little meaningful contribution to the expression of sarcasm.

\textbf{New Dataset and Baselines.}
To address the aforementioned issues, we propose a new dataset, \textbf{M2SaG}, consisting of 4,970 samples.
Each sample comprises an image, a sarcastic text, and an explicitly annotated sarcasm target, as illustrated in the left of Fig.~\ref{fig1}(b).
Compared to MuSG, our M2SaG dataset exhibits a higher mean sarcasm score of 0.7700 and a lower standard deviation of 0.1817, indicating the presence of stronger and more consistent sarcastic content.
We introduce ViSP, a benchmark model based on the pretrained Vision-and-Language Transformer (ViLT) \cite{vilt}, to evaluate M2SaG.
Specifically, our approach first employs the ViLT model to extract joint multimodal embeddings from image and text, which are then fed into BART \cite{bart} to generate sarcastic text.
Inspired by InstructGPT \cite{Instructgpt}, we develop a Proximal Policy Optimization (PPO)-based framework that incorporates a score-guided generation strategy.
In this framework, BART first generates multiple candidate sarcastic texts, which are then evaluated by DIP \cite{dip} to assign sarcasm scores that reflect the strength of sarcastic intent.
The scores are used as reward signals in the PPO loss to iteratively refine the generation process, steering the model toward outputs with stronger sarcastic intent.
Meanwhile, during training, BART generates multiple candidates, treating the highest-scoring text as the positive sample and the rest as negatives for contrastive learning, further enhancing the model’s ability to produce high-quality sarcasm.

To validate the effectiveness of ViSP, we conduct comprehensive comparisons with various baselines, including text-only models such as GPT-2 \cite{gpt2} and T5 \cite{t5},  Vision-Language Model (VLMs) such as BLIP \cite{blip} and GIT \cite{git}, and large language models (LLMs) such as LLaVA \cite{llava} and DeepSeek \cite{deepseek}.
ViSP consistently outperforms across all evaluation metrics, showcasing its strength in sarcasm generation.
Furthermore, we analyze the distributions of Sarcasm Scores and Factual Incongruity for both M2SaG and the texts generated by ViSP. 
The generated texts exhibit higher mean Sarcasm Scores (0.898 vs. 0.770) and Factual Incongruity (0.768 vs. 0.739), indicating that ViSP not only captures sarcastic intent more effectively but also introduces greater semantic contrast between image and text. 
These results demonstrate that ViSP produces sarcastic content of higher quality and stronger expressive clarity than the original dataset.
Detailed experiments and analysis are in Section Experiments.

The main contributions of this paper are summarized as follows:
\begin{itemize}
    \item We develop M2SaG, \textbf{a new dataset} comprising 4,970 samples, specifically designed for the task of multimodal sarcasm generation.
    \item We benchmark M2SaG using a novel encoder-decoder model built upon ViLT, serving as a strong baseline for multimodal sarcasm generation.
    \item To the best of our knowledge, we are the first to introduce reinforcement learning with PPO loss into the domain of sarcasm generation. Experimental results demonstrate its strong effectiveness in enhancing the quality of generated sarcastic text.
    \item We conduct comprehensive comparisons across text-only models, vision-language models (VLMs), and large language models (LLMs), revealing that LLMs underperform in sarcasm generation. These findings further highlight the effectiveness and robustness of ViSP.

\end{itemize}

\section{Realted Works}
\label{Realted Works} 
Existing research on sarcasm can be categorized into two main areas: sarcasm understanding, which includes sarcasm detection, explanation, and target identification, and sarcasm generation, which focuses on producing sarcastic text.

\noindent \textbf{Sarcasm detection.} 
Early studies used handcrafted features for sarcasm detection \cite{ealyscaram}. 
With social media’s growth, multimodal approaches to sarcasm understanding have gained increasing research interest.
\cite{mustard} collected 690 videos from YouTube to construct the MUStARD dataset, incorporating conversational context features to improve detection performance.
\cite{mmsd2} noted that the MMSD dataset \cite{MMSD} contains spurious cues, such as hashtags and emojis, and suffers from noisy labeling, where the absence of \textit{\#sarcasm} is incorrectly treated as non-sarcastic.
To address this, MMSD 2.0 provides cleaner, more reliable samples.
\cite{knowlenet, dip, dyn} pointed out that the semantic relationship between images and text in sarcastic content is often implicit or even contradictory. 
Existing vision-language models (VLMs), which are typically trained on semantically aligned image-text pairs, are ill-suited for the task of sarcasm detection.
To address this, \cite{knowlenet, g2sam, mdpf} incorporated external commonsense knowledge to enhance the understanding of implicit sarcastic meanings.
\cite{dip} captured sarcasm by modeling Factual and Affective Incongruity, identifying semantic conflict and emotional contrast in multimodal data to improve detection.
\cite{debiasing} revealed that multimodal sarcasm models often over-rely on biased textual cues and are sensitive to input perturbations. 
To mitigate this, a contrastive learning approach is proposed to reduce spurious dependencies while preserving semantic alignment.

\noindent \textbf{Sarcasm explanation.}
\cite{nice, did} argued that detecting sarcasm alone is insufficient for full comprehension, proposing the task of multimodal sarcasm explanation to uncover the underlying rationale.
\cite{nice} constructed the MORE dataset, a multimodal sarcasm explanation benchmark consisting of 3,510 sarcastic posts, each annotated with corresponding textual explanations to capture the underlying rationale behind the sarcastic intent.
\cite{team} noted that ExMORE \cite{nice} overlooks image object metadata and external knowledge, and thus integrate a knowledge graph via graph neural networks, achieving notable performance gains.
\cite{turbo} highlighted the importance of explicitly modeling sarcasm targets and extended the MORE dataset to MORE+ by adding target annotations to improve multimodal sarcasm analysis.
\cite{mose} proposed that sarcasm explanation facilitates a deeper understanding of affective tasks.
Consequently, they trained a sarcasm explanation model designed to jointly address sarcasm detection, humor recognition, and sentiment analysis.
\cite{ccg} argued that prior methods neglect cross-modal attention.
They proposed a context-aware embedding framework to better fuse multimodal information.

\noindent \textbf{Sarcasm target identification.}
\cite{msti} introduced a novel task, Multimodal Sarcasm Target Identification (MSTI), which fully leveraged both visual and textual information to identify the specific target of sarcasm within multimodal content.
\cite{cofipara} argued that sarcasm detection was a coarse-grained task and that deeper investigation and understanding were necessary.
Their approach first pre-trained a model on coarse-grained sarcasm detection and subsequently fine-tuned it on the more fine-grained MSTI task.

\noindent \textbf{Sarcasm generation.}
\cite{mtmsg} developed a multimodal sarcasm generation dataset, MuSG, comprising 5,000 samples.
They further proposed a baseline model, MTMSG, which leveraged both OCR text and visual information, with the OCR text typically providing contextual cues essential for interpreting sarcasm.

\section{Dataset}
\label{dataset}
This section provides a detailed introduction to the \textbf{M}ulti\textbf{m}odal \textbf{Sa}rcasm \textbf{G}eneration (M2SaG) dataset we develop. 
An illustrative example is presented in Fig.~\ref{fig1}(b).
Given the weak image-text alignment and vague sarcasm targets in MuSG \cite{mtmsg}, we construct a new dataset that ensures clear sarcasm target annotations and strong visual-textual alignment.
We explore two existing datasets, MSTI \cite{msti} and MORE+ \cite{turbo}, to obtain sarcasm targets.
The MSTI dataset\footnote{https://github.com/Lbotirx/CofiPara/blob/master/data/msti} contains 10,506 image-text pairs, each annotated with the specific target of sarcasm in both modalities, tailored for multimodal sarcasm target identification.
The MORE+ dataset\footnote{https://github.com/flamenlp/TURBO/tree/main} extends the original MORE dataset by adding explicit sarcasm target annotations to 3,510 samples, better capturing sarcasm directed at specific targets \cite{turbo}.

\textbf{Exclusion.}
To construct a high-quality multimodal sarcasm generation dataset, we apply the following filtering criteria to the aforementioned two datasets, as follows:
\begin{itemize}
    \item[\textendash] Samples with sarcastic texts exceeding 40 words are discarded.
    \item[\textendash] Samples with extraneous quotation marks at the beginning and end of the text, as well as spaces preceding punctuation marks, are discarded.
    \item[\textendash] Samples with sarcasm scores lower than 0.5 are discarded.
\end{itemize}

Specifically, the MSTI dataset annotates sarcasm targets within the text using the BIO tagging scheme, where ``B-S'' denotes the beginning of a sarcasm target and ``I-S'' indicates its continuation.
We select samples containing the annotations ``B-S'' and ``I-S'' simultaneously and extract the corresponding sarcasm targets from these annotations, yielding a total of 3,419 samples.
Subsequently, We filter out samples with sarcastic texts longer than 40 words and clean the texts by removing leading/trailing quotation marks and extra spaces before punctuation, resulting in 3,385 refined samples.
Furthermore, we employ the pretrained DIP \cite{dip} to evaluate each sample, removing those with scores below 0.5, which results in a subset of 2,678 samples.
For the MORE+ dataset, we similarly remove sarcastic texts over 40 words and clean extraneous quotes and spaces before punctuation, yielding 3,512 samples.
We evaluate the remaining samples with DIP, filtering out those scoring below 0.5, which results in 2,279 samples. 
Combined with other data, this yields our final M2SaG dataset of 4,970 samples.
The entire dataset is randomly shuffled and partitioned into training, validation, and test sets with an 8:1:1 ratio, resulting in 3,976, 497, and 497 samples respectively.
A brief statistical summary of the dataset is presented in Table ~\ref{tab1}.
\begin{table}[t]
\centering
\setlength{\tabcolsep}{1mm}
\begin{tabular}{lccccc}
    \toprule
    \textbf{Split} & \textbf{\# of Posts} & \multicolumn{2}{c}{\textbf{Sarcastic text}} & \multicolumn{2}{c}{\textbf{Sarcasm target}} \\
    \cmidrule(lr){3-4} \cmidrule(lr){5-6}
     &  & \textbf{Avg. length} & \textbf{$|V|$} & \textbf{Avg. length} & \textbf{$|V|$} \\
    \midrule
    Train & 3,976 & 16.39 & 9,927 & 2.72 & 3,608 \\
    Val & 497 & 16.25 & 2,575 & 2.69 & 778 \\
    Test & 497 & 17.38 & 2,570 & 2.65 & 747 \\
    \bottomrule
\end{tabular}
\caption{Statistics of the M2SaG dataset.}
\label{tab1}
\end{table}

\begin{figure*}[t]
\centering
    \includegraphics[width=0.75\textwidth]{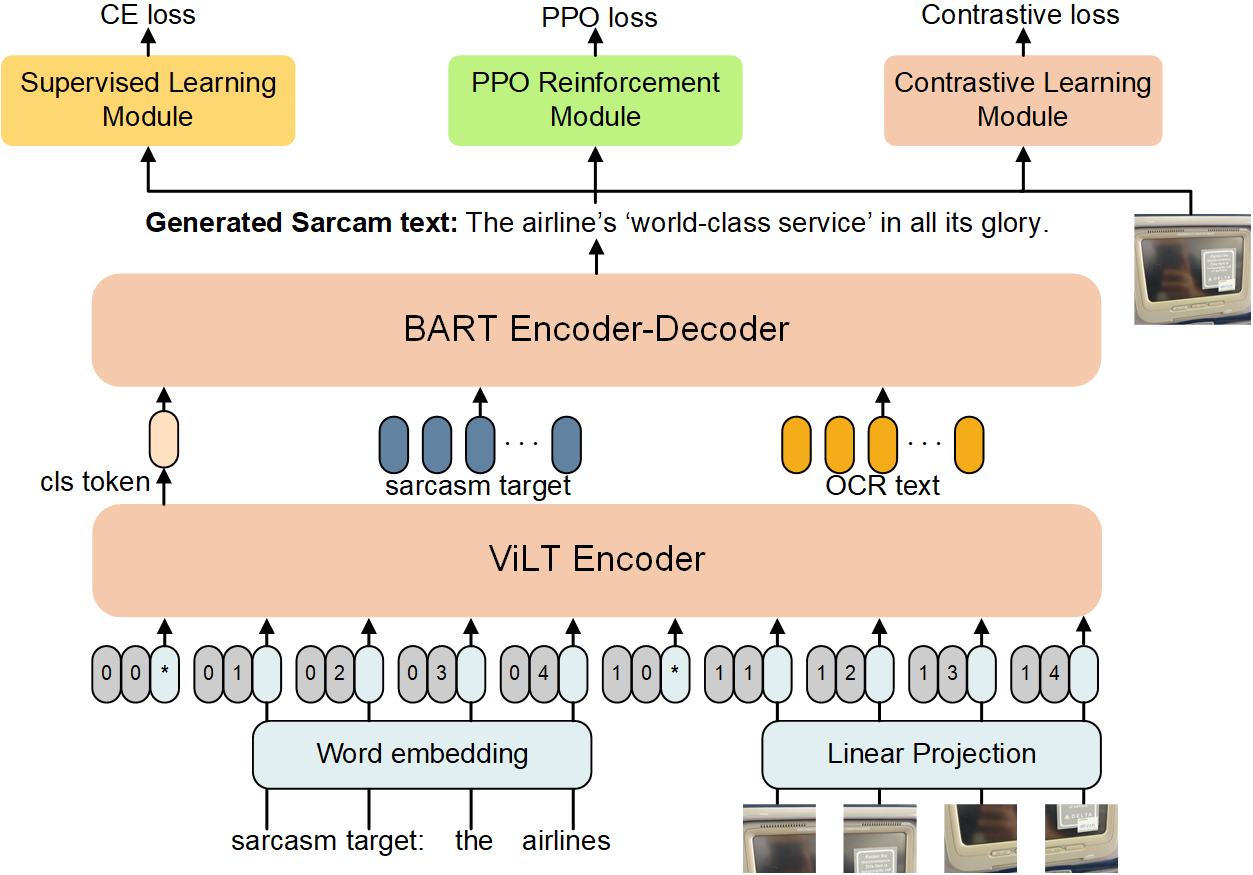} 
    \caption{The overall architecture of ViSP.}
\label{fig2}
\end{figure*}

\section{Methodology}
\label{Methodology}
To generate sarcastic texts, we propose ViSP, an encoder-decoder framework based on ViLT. 
This section provides a detailed description of the overall architecture of ViSP and its individual components.

\subsection{Overview of our proposed method}
\label{overview}
Fig.~\ref{fig2} illustrates the overall architecture of ViSP, which consists of five key modules: the Multimodal Encoding Module, the Generation Module, the Supervised Learning Module, the PPO Reinforcement Module, and the Contrastive Learning Module.
Specifically, given a sample (\textit{I, T, ST}), where \textit{I} denotes the image, \textit{T} the sarcastic text, and \textit{ST} the sarcasm target, the OCR text ($OT$) is first extracted from \textit{I} using EasyOCR. 
We concatenate $ST$ and $OT$ into a sequence $T'$, which, along with $I$, is input to ViLT for a fused multimodal representation.
The fused [CLS] token and token-level embeddings of $T'$ from ViLT are concatenated and fed into the BART encoder. 
The BART decoder then generates the sarcastic text $T$.
During training, the model generates top-$k$ candidate texts evaluated by DIP.
The highest-scoring candidate is treated as positive, and others as negatives for contrastive learning. 
The top candidate also guides policy optimization via PPO for reward-driven learning.
At inference, the model generates top-$k$ candidates and selects the highest-scoring one as the final output.

\subsection{Multimodal Encoding Module}
\label{mme}
To guide the generation process, we construct a prompt template for the sarcasm target: “\textit{The target of sarcasm is \{\}. Write a sarcastic comment based on this.}”
Subsequently, we examine the effect of three input components on sarcasm generation: OCR text, image caption, and image object.

\textbf{OCR text.}
We argue that the OCR text embedded in images is often closely related to the target of sarcasm.
For example, in Fig.~\ref{fig1}(b), the sarcasm target “7:22 7w not coming” explicitly contains “7w,” which is directly obtained through OCR.
Therefore, we utilize EasyOCR to extract OCR content from the images.
The extracted text is then converted into a string using the template format: \textit{OCR text: \{\}}.

\textbf{Image caption.}
In addition to OCR text, we consider image captions to be valuable for providing contextual and object-level descriptions of the visual scene, which can facilitate the generation of more coherent and context-aware sarcastic texts. 
To this end, we use BLIP \cite{blip} to generate image captions that capture salient visual elements and relationships.
Similarly, the generated image captions are converted into string format using the template: \textit{Image caption: \{\}}.

\textbf{Image object.}
Sarcasm is often directed toward a specific object present in either the image or the accompanying text. 
Explicitly providing the model with such potential sarcasm targets can facilitate more accurate and contextually appropriate generation.
To this end, we utilize the YOLOv11 \cite{yolo11} to detect objects in each image and eliminate duplicates.
The resulting object list is then converted into a textual string using the template: \textit{Objects in image: \{\}}.

After constructing the complete textual input, denoted as $T'$, we feed both $I$ and $T'$ into a pretrained ViLT encoder to extract the fused multimodal embeddings.

\subsection{Generation Module and Supervised Learning Module}
\label{generate}
\textbf{Generation Module.}
Following \cite{nice,team}, we also employ BART for text generation.
Given that BART is pretrained exclusively on textual inputs, we adopt a lightweight fusion strategy by leveraging only the [CLS] token from ViLT as the multimodal representation. 
This token is concatenated with $T'$ and fed into BART, which is subsequently fine-tuned to generate the sarcastic text $T$.

\textbf{Supervised Learning Module.}
During training, to enhance generation quality, we prompt BART to produce top-k candidate sarcastic texts for each input. 
Among these, the highest-scoring candidate is selected as it best aligns with the image.
We then compute the cross-entropy loss between this selected candidate and the ground-truth sarcastic text, denoted as $\mathcal{L}_{\text{ce}}$, which is formulated as follows:

\begin{equation}
    \mathcal{L}_\text{ce} = -\sum_{t=1}^{T} \log P_\theta \bigl(y_t^* \mid y_{<t}^*, \mathbf{z}\bigr)
\label{ce loss}
\end{equation}

where $y_t^*$ denotes the $t$-th token of the ground truth sequence, and $\mathbf{z}$ represents the [CLS] token output from the ViLT encoder.

\subsection{Contrastive Learning Module}
\label{cl}

\begin{figure}[t]
\centering
	{\includegraphics[width = 0.5\textwidth]{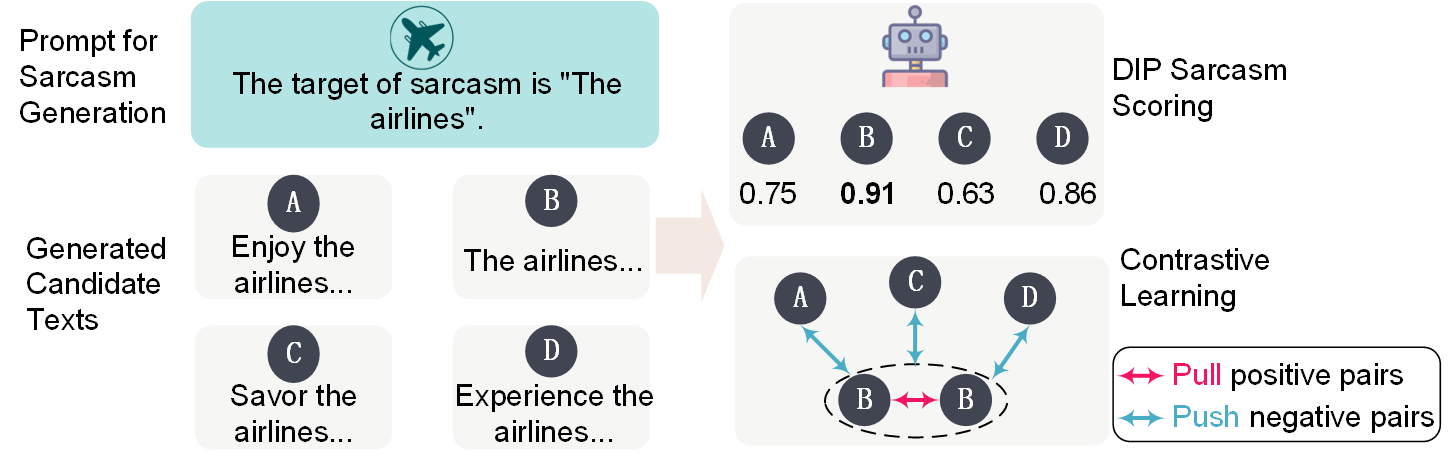}}
\caption{Contrastive Learning Module. We designate the highest-scoring generated text as a pseudo ground-truth anchor, while the remaining lower-scoring candidates are treated as negative samples.}
\label{fig3}
\end{figure}

Generating a single sarcastic text often fails to capture the full complexity and diversity inherent in sarcasm, limiting the variety and quality of the output.
To address this issue, we propose a Contrastive Learning Module.
In contrastive learning module, as illustrated in Fig.~\ref{fig3}, we first generate the top-$k$ candidate texts for each input and evaluate their sarcasm quality using DIP.
The candidate with the highest score is selected as a pseudo ground-truth anchor, while the remaining lower-scoring candidates are treated as negative samples. 
In Fig.~\ref{fig3}, for example, text B serves as the anchor, while A, C, and D are considered negative samples.
The contrastive learning loss function $\mathcal{L}_{cl}$ is defined as follows:

\begin{equation}
\mathcal{L}_\text{cl} = -\frac{1}{N} \sum_{i=1}^{N} \log \frac{\exp\bigl(\text{sim}(\mathbf{g}_i, \mathbf{g}_i^+)/\tau\bigr)}{\sum_{j=1}^{N-1} \exp\bigl(\text{sim}(\mathbf{g}_i, \mathbf{g}_j^-)/\tau\bigr)}
\label{cl loss}
\end{equation}

where $\mathbf{g}_i$ denotes the embedding of the anchor (i.e., the highest-scoring generated text), $\mathbf{g}_i^+$ denotes the embedding of the positive sample (identical to the anchor), and $\mathbf{g}_i^-$ are the embeddings of the $N-1$ negative samples.
This contrastive strategy encourages the model to better distinguish semantic quality, enhancing its ability to generate high-quality sarcastic texts.

\begin{table*}[htbp]
    \centering
    \normalsize
    \begin{tabular}{l|l|c|c|c|c|c|c|c|c|c|c}
    \hline
    \multirow{2}{*}{} & \multirow{2}{*}{Model} & \multicolumn{4}{c|}{BLEU} & \multicolumn{3}{c|}{Rouge} & \multirow{2}{*}{Cider} & \multirow{2}{*}{METEOR} & \multirow{2}{*}{Sent-BERT} \\ \cline{3-9}
     &  & B1 & B2 & B3 & B4 & RL & R1 & R2 &  &  &  \\ \hline
    \multirow{3}{*}{Text-only} & GPT2 & 10.39 & 3.66 & 1.73 & 0.93 & 11.31 & 16.13 & 4.43 & 5.58 & 15.96 & 64.66 \\ 
     & OPT & 10.33 & 3.82 & 1.87 & 1.04 & 10.96 & 18.14 & 4.11 & 7.76 & 16.89 & 65.06 \\ 
     & GPT-Neo & 10.06 & 3.79 & 1.93 & 1.13 & 9.04 & 16.09 & 3.66 & 7.27 & 16.09 & 65.79 \\ 
     & T5 & 12.73 & 5.31 & 2.79 & 1.57 & 13.01 & 16.58 & 3.45 & 11.41 & 16.70 & 73.57 \\ \hline
    \multirow{3}{*}{LLMs} & LLaVA & 7.99 & 3.48 & 1.85 & 1.14 & 9.36 & 15.09 & 3.58 & 10.68 & 13.35 & 69.69 \\ 
     & DeepSeek & 4.02 & 1.19 & 0.52 & 0.28 & 4.64 & 8.01 & 0.97 & 0.45 & 8.60 & 62.36 \\ 
     & Qwen & 5.09 & 1.71 & 0.79 & 0.41 & 5.77 & 11.49 & 1.99 & 0.26 & 11.29 & 65.86 \\ 
     & Hunyuan & 5.13 & 1.59 & 0.62 & 0.24 & 3.43 & 8.96 & 1.40 & 1.34 & 8.18 & 46.53 \\ \hline
    \multirow{3}{*}{VLMs} & ViT-GPT2 & 4.87 & 0.62 & 0.11 & 0.04 & 7.21 & 6.02 & 0.03 & 0.03 & 10.88 & 61.80 \\ 
     & MIC & 14.08 & 6.11 & 2.84 & 0.96 & 12.24 & 8.79 & 0.71 & 4.02 & 9.68 & 60.12 \\ 
     & GTT & 13.10 & 5.25 & 2.55 & 1.29 & 19.84 & 16.57 & 4.49 & 0.46 & 17.53 & 64.22 \\ 
     & \textbf{ViSP(ours)} & \textbf{24.26} & \textbf{13.71} & \textbf{8.92} & \textbf{5.68} & \textbf{21.50} & \textbf{18.35} & \textbf{4.81} & \textbf{35.71} & \textbf{17.79} & \textbf{75.08} \\ \hline
    \end{tabular}
    \caption{Comparative analysis on the M2SaG dataset. ViSP achieves the best results across all five sets of evaluation metrics.}
    \label{tab2}
\end{table*}

\subsection{PPO Reinforcement Module}
\label{ppo Module}

Inspired by InstructGPT \cite{Instructgpt}, we propose leveraging Proximal Policy Optimization (PPO) reinforcement learning to encourage the model to generate higher-quality sarcastic texts that better align with the input images.

\textbf{Reward model.}
We adopt DIP as a reward model to optimize the generation policy. 
Specifically, DIP serves as a sarcasm-aware evaluator that provides feedback on the quality of generated texts in terms of their sarcastic intent.
Leveraging DIP as a reward signal helps guide the model to produce more sarcastic and contextually appropriate texts.

\textbf{Supervised Fine-Tuning policy.}
To stabilize policy updates and avoid divergence from the pretrained behavior, we use a momentum encoder similar to that in MoCo \cite{moco} to approximate the Supervised Fine-Tuning (SFT) policy $\pi^{\text{SFT}}$.
Specifically, we initialize the momentum encoder $\pi^\mathrm{SFT}$ with the policy network's parameters and update it using an exponential moving average (EMA) of the current reinforcement learning policy $\pi^\mathrm{RL}$.
The update rule is defined as:

\begin{equation}
\theta_t^\text{SFT} = m \cdot \theta_{t-1}^\text{SFT} + (1 - m) \cdot \theta_t^\text{RL}
\end{equation}

Where $\theta_t^\text{SFT}$ and $\theta_t^\text{RL}$ denote the parameters of the momentum encoder and RL policy at step $t$, respectively, with $m \in [0, 1]$ as the momentum coefficient.

\textbf{KL Divergence Penalty.}
To ensure stable policy updates, PPO introduces a KL divergence penalty that constrains the deviation between the current policy $\pi^\mathrm{RL}$ and the reference policy $\pi^\mathrm{SFT}$.


The PPO loss is formulated as shown in Eq.~\ref{ppo}:

\begin{equation}
    \mathcal{L}_{\text{ppo}} = \mathbb{E}_{(x, y) \sim \pi^{\text{RL}}_\phi} \left[ r_\theta(x, y) - \beta \log \left( \frac{\pi^{\text{RL}}_\phi(y \mid x)}{\pi^{\text{SFT}}(y \mid x)} \right) \right]
\label{ppo}
\end{equation}

Where $r_\theta(x, y)$ denotes the reward score from DIP, and $\beta$ controls the KL penalty strength.

\subsection{Overall Loss}

We combine all the loss functions to obtain the final loss function $\mathcal{L}$, as shown in Eq.~\ref{loss}
\begin{equation}
\mathcal{L} = \mathcal{L}_{ce} +\lambda_{ppo}\mathcal{L}_{ppo}+\lambda_{cl}\mathcal{L}_{cl}
\label{loss}
\end{equation}

\section{Experiments}
\label{experiments}

\subsection{Experiment Settings}
All experiments are conducted on a single NVIDIA A100-PCIE-40G GPU.
We train the model for 20 epochs with a batch size of 16 and an initial learning rate of 1e-4, including the first 100 steps for warm-up.
We use the BART tokenizer with a maximum token length of 256.
The contrastive loss weight $\lambda_{\text{cl}}$ is fixed at 0.5, while the PPO loss weight $\lambda_{\text{ppo}}$ varies dynamically during training.
The temperature parameter $\tau$ in $\mathcal{L}_{\text{ppo}}$ is set to 0.07. 
Additionally, the model generates top-$k$ candidates (with $k=5$) during both training and inference. 
Detailed explanations of these hyperparameters are provided in Section Ablations.

\subsection{Comparison Models}
We compare ViSP against the following three categories of models. 
All models are obtained from the HuggingFace\footnote{https://huggingface.co/} repository.

\textbf{(1) Text-Only Models:} 
GPT2 \cite{gpt2}, OPT-125M \cite{opt}, GPT-Neo-125M \cite{gptneo}, T5-base \cite{t5}.

\textbf{(2) Large Language Models (LLMs):} 
LLaVA-1.5-7B \cite{llava}, DeepSeek-R1-0528-Qwen3-8B \cite{deepseek}, Qwen3-8B \cite{qwen3}, Hunyuan-7B-Pretrain \cite{hunyuan}.

\textbf{(3) Vision-Language Models (VLMs):} 
VIT-GPT2 \cite{vit-gpt2}, Mini-image-captioning, GIT-base \cite{git}.

\begin{table*}[htbp]
\setlength{\tabcolsep}{1mm}
    \centering
    \begin{subtable}[t]{0.3\textwidth}
        \centering
        \begin{tabular}{lccc}
            \toprule
            Case & B1 & METEOR \\
            \toprule
            st & 16.03 & 15.14  \\ 
            \hspace{1em}- ocr & \cellcolor{gray!25} 21.32 & \cellcolor{gray!25} \textbf{18.44}  \\ 
            \hspace{1em}- obj & 18.66 & 16.54  \\ 
            \hspace{1em}- desc & 20.04 & 16.09 \\ 
            \hspace{1em}- ocr + obj & 21.10 & 16.53  \\ 
            \hspace{1em}- ocr + desc & \textbf{21.36} & 17.64  \\ 
            \hspace{1em}- ocr + obj + desc & 20.65 & 17.64  \\ 
            \bottomrule
        \end{tabular}
        \label{ablation_a}
        \caption{Textual Input}
    \end{subtable}
    \hfill
    \begin{subtable}[t]{0.3\textwidth}
        \centering
        \begin{tabular}{lccc}
            \toprule
            Case & B1 & METEOR \\
            \toprule
            ce & 19.95 & 16.73  \\ 
            \hspace{1em}- ppo &  22.92 &  \textbf{16.92} \\ 
            \hspace{1em}- cl & 22.53 & 16.16  \\ 
            \hspace{1em}- ppo + cl &  \cellcolor{gray!25} \textbf{22.93} & \cellcolor{gray!25} 16.89 \\ 
            \bottomrule
             &  &  &  \\
             &  &  &  \\
             &  &  &  \\
        \end{tabular}
        \caption{Loss fuction}
    \end{subtable}
    \hfill
    \begin{subtable}[t]{0.3\textwidth}
        \centering
        \begin{tabular}{cccc}
            \toprule
            Train & Inference & B1 & METEOR \\
            \toprule
            1 & 1 & 23.04 & 16.99  \\ 
            5 & 1 & 22.93 & 16.89 \\ 
            5 & 5 &  \cellcolor{gray!25}  \textbf{24.26} &  \cellcolor{gray!25}  \textbf{17.79 } \\
            \bottomrule
             &  &  &  \\
             &  &  &  \\
             &  &  &  \\
             &  &  &  \\
        \end{tabular}
        \caption{Number of generated texts}
    \end{subtable}

     \begin{subtable}[t]{0.3\textwidth}
        \centering
        \begin{tabular}{lccc}
            \toprule
             $\lambda_{\text{cl}}$ & B1 & METEOR \\
            \toprule
            0.05 & 23.84 & 17.88 \\ 
            0.1 & 22.64 &  16.59\\ 
            0.5 &  \cellcolor{gray!25} \textbf{24.26} &  \cellcolor{gray!25} \textbf{17.79} \\ 
            1.0 & 22.92 & 17.16 \\  
            \bottomrule
        \end{tabular}
        \caption{$\lambda_{ \text{cl}}$}
    \end{subtable}
    \hfill
    \begin{subtable}[t]{0.3\textwidth}
    \centering
        \begin{tabular}{lccc}
            \toprule
            Case & B1 & METEOR \\
            \toprule
            w/ $\lambda_{\text{ppo}}$ & \cellcolor{gray!25} \textbf{23.47} & \cellcolor{gray!25} \textbf{17.53}\\ 
            w/o $\lambda_{\text{ppo}}$ & 22.43 & 16.77\\  
            \bottomrule
            &  &  &  \\
             &  &  &  \\
        \end{tabular}
        \caption{$\lambda_{ \text{ppo}}$}
    \end{subtable}
    \hfill
     \begin{subtable}[t]{0.3\textwidth}
        \centering
        \begin{tabular}{lccc}
            \toprule
             Case & B1 & METEOR \\
            \toprule
            w/ [cls] & \cellcolor{gray!25} \textbf{24.26} & \cellcolor{gray!25} \textbf{17.79} \\ 
            w/o [cls] & 23.99 & 17.71\\  
            \bottomrule
            &  &  &  \\
             &  &  &  \\
        \end{tabular}
        \caption{Multimodal embedding [CLS] token}
    \end{subtable}
    \caption{ViSP ablation experiments. We report BLUE-1 (B1) and METEOR accuracy (\%). Default settings are marked in gray.}
\label{ablation table}
\end{table*}

\subsection{Main Results}
Table~\ref{tab2} presents the comparative results on the M2SaG dataset.
Among text-only models, T5 performs best on most metrics, except ROUGE-1, ROUGE-2, and METEOR.
Within LLMs, LLaVA achieves the highest scores across all metrics, yet still lags behind multimodal methods.
ViSP consistently outperforms all baselines across five evaluation sets, achieving BLEU-1 to BLEU-4 scores of 24.26 (+10.18), 13.71 (+7.60), 8.92 (+6.08), and 5.68 (+4.11), respectively.
It also obtains ROUGE-L of 21.5 (+1.66), ROUGE-1 of 18.35 (+0.21), ROUGE-2 of 4.81 (+0.32), Cider of 35.71 (+24.3), METEOR of 17.79 (+0.26), and Sent-BERT of 75.08 (+5.39).
Notably, LLMs exhibit the weakest performance overall, underscoring the challenges they face in capturing sarcasm effectively.

\begin{figure*}[ht]
\centering
	\subfloat[Sarcasm Scores]{\includegraphics[width = 0.5\textwidth]{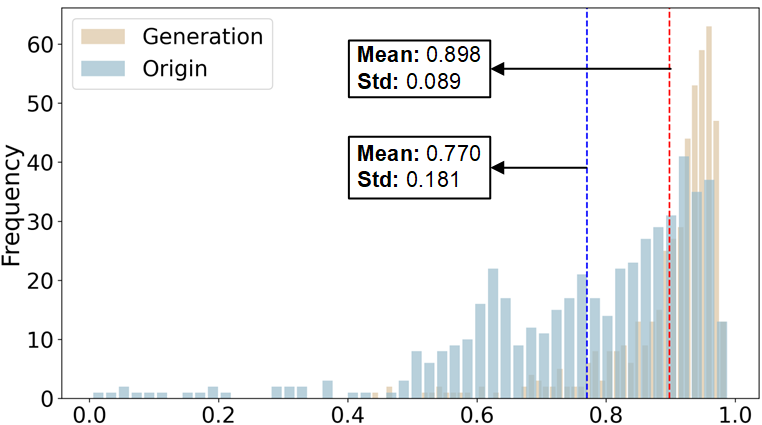}}
	\subfloat[Factual Incongruity]{\includegraphics[width = 0.5\textwidth]{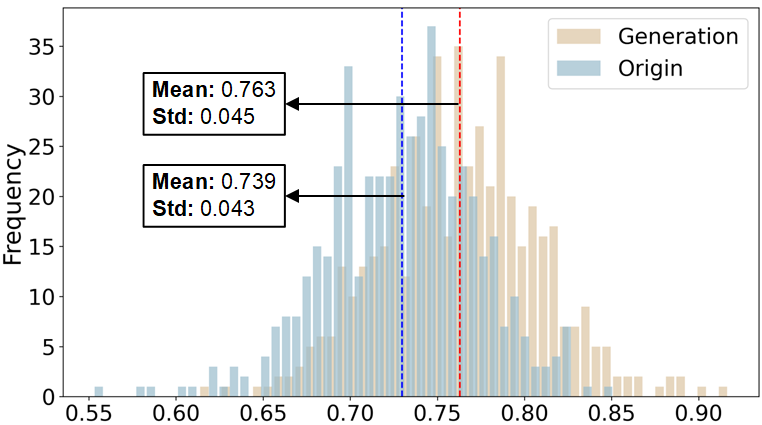}}
\caption{Evaluation of generated sarcastic texts using Sarcasm Scores and Factual Incongruity.}
\label{fig4}
\end{figure*}

\subsection{Ablations}
\label{ablation}
We conduct ablation studies on the individual components of ViSP to validate the effectiveness of each component.
In the ablation experiments, we report BLEU-1 (B1) and METEOR as the evaluation metrics, as shown in Table~\ref{ablation table}.

\textbf{Textual Input.}
To identify the optimal input configuration, we start with $\mathcal{L}_{\text{ce}}$ and evaluate the impact of each component: sarcasm target (\textbf{st}), image object (\textbf{obj}), and image description (\textbf{desc}). 
As shown in Table~\ref{ablation table}(a), using only \textbf{st} yields suboptimal results.
Adding OCR text improves BLEU-1 by 5.29\% and METEOR by 3.3\%, as it often directly contributes to the sarcasm. 
In contrast, incorporating \textbf{obj} and \textbf{desc} provides limited gains, likely due to their generic or redundant nature. 
Unless otherwise stated, we use \textbf{st} and OCR text as default inputs.

\textbf{Loss Fuction.}
To assess the effect of each loss, we set all weights to 1 (Table~\ref{ablation table}(b)) and generate a single output at inference.
Using $\mathcal{L}_{\text{ce}}$ alone yields a BLEU-1 of 19.95\%.
Incorporating $\mathcal{L}_{\text{ppo}}$ brings a 2.97\% gain in BLEU-1 and a slight 0.19\% improvement in METEOR.

\textbf{Number of generated texts.}
Since generating a single sarcastic text often fails to capture the diversity of sarcasm, we explore generating top-$k$ candidates ($k=5$) during training and inference. 
As shown in Table~\ref{ablation table}(c), this yields improved performance, with BLEU-1 and METEOR increasing by 1.22\% and 0.8\%, respectively.

\textbf{$\lambda_{ \text{ppo}}$.}
Given the potential instability of $\mathcal{L}_{\text{ppo}}$ early in training due to conflict with supervised objectives, we perform an ablation on its weight $\lambda_{\text{ppo}}$ using only $\mathcal{L}_{\text{ce}}$ and $\mathcal{L}_{\text{ppo}}$. 
A linear schedule is adopted, gradually increasing $\lambda\_{\text{ppo}}$ to smoothly introduce reinforcement learning into training.
As shown in Table~\ref{ablation table}(e), this strategy proves effective.

\textbf{$\lambda_{ \text{cl}}$.}
We also investigate the impact of $\lambda_{\text{cl}}$. 
When $\lambda_{\text{cl}} = 0.5$, the BLEU-1 and METEOR scores reach 24.26\% and 17.53\%, respectively, as shown in Table~\ref{ablation table}(d).

\textbf{Multimodal embedding [CLS] token.}
We examine the [CLS] token from ViLT, which encodes global image-text semantics.
As shown in Table~\ref{ablation table}(f), incorporating it yields BLEU-1 and METEOR gains of +0.27\% and +0.08\%, confirming the benefit of global multimodal information for sarcasm generation.

\subsection{Analysis of Generated Sarcastic Texts}
\label{Analysis}
As shown in Fig.~\ref{fig4}, we compute sarcasm scores and factual incongruity on the test set and generated texts to assess sarcasm quality.

\textbf{Sarcasm scores.}
We compute the sarcasm scores of generated texts and compare them with annotated sarcastic texts in Fig.~\ref{fig4}(a).
Compared to the original data, the generated texts exhibit a higher mean sarcasm score of 0.898 and a lower variance of 0.089, indicating not only an overall improvement in sarcastic expression but also more consistent sarcasm generation. 

\textbf{Factual Incongruity.}
Researchers have observed that sarcasm often arises when there is a discrepancy between the literal meaning and the observed facts, a process referred to as counterfactual inference \cite{counterfactual}.
Following DIP \cite{dip}, we compute the image-text similarity score $S_{inter}$ using CLIP, and define factual incongruity as $1- S_{inter}$.
As shown in Fig.~\ref{fig4}(b), the generated sarcastic texts exhibit a higher factual incongruity, with a mean of 0.763 compared to 0.739 for the original data, indicating a stronger degree of sarcasm.

\section{Discussion and Conclusion}
\label{conclusion}
In this work, we introduce a novel dataset, M2SaG, consisting of 4,970 samples. 
Furthermore, we propose a robust baseline model, ViSP, to benchmark the M2SaG dataset.
To the best of our knowledge, this is the first study to incorporate reinforcement learning via PPO into the sarcasm generation domain.
Our experiments demonstrate the best performance across five evaluation metrics. 
We also show that large language models (LLMs) perform suboptimally on sarcasm generation.
Additionally, we conduct extensive analyses on the generated sarcastic texts.
However, ViSP still has the following limitations:
(1) Model performance relies on the quality of the external sarcasm evaluator and poor evaluators may introduce bias.
(2) PPO loss is sensitive and unstable, often causing training oscillations and slow convergence.
(3) Naively concatenating OCR text with sarcasm targets may dilute key information and introduce noise.

In future work, we plan to:
(1) jointly train the sarcasm generator and evaluator in an adversarial manner, inspired by GANs, where the evaluator provides feedback to refine generation quality;
(2) explore reinforcement learning methods with more sophisticated reward designs to improve both stability and performance;
(3) investigate prompt-based instructional guidance to inject task-specific knowledge and enhance the model’s control over sarcastic text generation.



\bibliography{aaai2026}

\end{document}